\documentclass{article}

\usepackage[preprint]{neurips_2025}

\workshoptitle{MATH-AI}

\usepackage[utf8]{inputenc} 
\usepackage[T1]{fontenc}    
\usepackage[colorlinks=true]{hyperref}
\hypersetup{
    linkcolor=cyan,       
    citecolor=cyan,      
    filecolor=cyan,    
    urlcolor=cyan         
}
\usepackage{url}            
\usepackage{booktabs}       
\usepackage{amsfonts}       
\usepackage{nicefrac}       
\usepackage{microtype}      
\usepackage{xcolor}         
\usepackage{pifont} 
\usepackage{wrapfig,graphicx}
\title{Max It or Miss It: Benchmarking LLM \\On Solving Extremal Problems}

\author{%
  Binxin Gao$^{1,\dagger}$\quad Jingjun Han$^2$ \\
  $^1$University of Maryland\quad $^2$Fudan University\\
  $^\dagger$ Corresponding author \\
  \texttt{bgao666@umd.edu}
}

\begin{document}

\maketitle

\begin{abstract}
Test-time scaling has enabled Large Language Models~(LLMs) with remarkable reasoning capabilities, particularly in mathematical domains, through intermediate chain-of-thought~(CoT) reasoning before generating final answers. However, the specific sources and mechanisms underlying these reasoning capabilities remain insufficiently understood. Optimization reasoning, \textit{i.e.} finding extrema under constraints, represents a fundamental abstraction that underpins critical applications in planning, control, resource allocation, and prompt search. To systematically evaluate this capability, we introduce \texttt{\textbf{ExtremBench}}, a benchmark dataset for solving mathematical extremal problems, curated from inequality exercises used for Chinese Mathematical Olympiad and transformed into $93$ standardized extrema-finding problems. We conduct extensive evaluations across various state-of-the-art open-source model families, including the Qwen3, GPT-OSS, and DeepSeek. Our results reveal that LLMs' extremal-solving reasoning capabilities do not always align with those of current mathematical benchmarks such as AIME25 and MATH-500, with some models showing strong general mathematical reasoning but poor extremal-solving skills, and vice versa. This discrepancy highlights a critical gap in current evaluation practices and suggests that existing benchmarks may not comprehensively capture the full spectrum of mathematical reasoning abilities.
\footnote{Our benchmark dataset is available at \href{https://huggingface.co/datasets/binxingao/extrem-bench}{https://huggingface.co/datasets/binxingao/extrem-bench}.}
\end{abstract}

\section{Introduction}

\begin{wrapfigure}{r}{0.40\textwidth}
    \centering
    \vspace{-22pt}
    \includegraphics[width=0.40\textwidth]{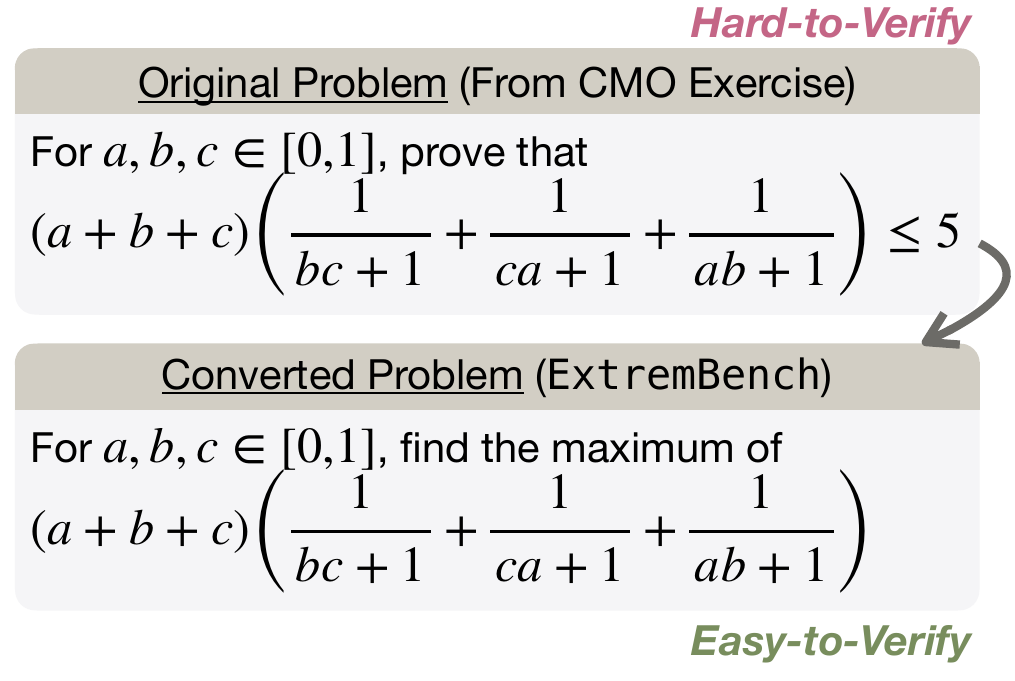}
    \vspace{-22pt}
    \caption{\texttt{ExtremBench} turns proof-style inequalities into equivalent extrema tasks, preserving reasoning challenge but enabling automatic evaluation.}
    \vspace{-18pt}
    \label{fig:teaser}
\end{wrapfigure}
Test-time scaling has enabled Large Language Models~(LLMs) with remarkable reasoning capabilities, particularly in mathematical domains, through intermediate chain-of-thought~(CoT) reasoning before generating final answers \citep{deepseek2025r1,openai2024openaio1card, snell2024scaling, yang2025qwen3}. However, the specific sources and mechanisms underlying these reasoning capabilities remain insufficiently understood. Optimization reasoning, \textit{i.e.}, finding extrema under constraints, represents a fundamental abstraction that underpins critical applications in planning, control, resource allocation, and prompt search~\citep{Wang_2024,zou2025llmbasedhumanagentcollaborationinteraction}. From determining optimal hyperparameters in machine learning~\citep{jin2024conditionalloraparametergeneration} to solving economic equilibrium problems~\citep{karten2025llm}, the ability to identify extrema under given constraints forms a cornerstone of mathematical problem-solving.

Despite the fundamental importance of extremal-solving capabilities, current mathematical benchmarks such as GSM8K \citep{cobbe2021training}, MATH-500 \citep{hendrycks2021measuring}, and AIME have barely evaluated this specific form of reasoning. These benchmarks primarily focus on algebraic manipulation and arithmetic computation, leaving optimization reasoning largely unexplored. This gap is particularly concerning given that extremal problems require a distinct set of reasoning skills including identifying constraint boundaries, understanding trade-offs between competing objectives, and recognizing when optimal solutions occur at critical points or boundaries.

\textit{Can we create a math benchmark that specifically evaluates LLM's extremal reasoning capabilities?} To systematically evaluate it, we introduce \texttt{ExtremBench}, a benchmark dataset for solving mathematical extremal problems, curated from inequality exercises used in Chinese Mathematical Olympiad and transformed into standardized extrema-finding tasks. Specifically, our transformation proces converts problems like ``\textit{For positive reals $a, b, c$ with $a + b + c = 3$, prove that $\frac{1}{a} + \frac{1}{b} + \frac{1}{c} \geq 3$}" into ``\textit{Find the minimum value of $\frac{1}{a} + \frac{1}{b} + \frac{1}{c}$ subject to $a, b, c > 0$ and $a + b + c = 3$.}" This reformulation, as demonstrated in Figure~\ref{fig:teaser}, preserves the mathematical complexity and reasoning requirements while creating a standardized and easy-to-verify~\footnote{We use \href{https://github.com/huggingface/Math-Verify}{https://github.com/huggingface/Math-Verify} for answer verification.} format for evaluating optimization capabilities.

We conduct extensive evaluations across various state-of-the-art open-source reasoning LLM families, including Qwen3~\citep{yang2025qwen3}, GPT-OSS~\citep{openai2025gptoss120bgptoss20bmodel}, and DeepSeek~\citep{deepseek2025r1}. Our results reveal that LLMs' extremal-solving reasoning capabilities do not always align with their performance on current mathematical benchmarks such as AIME25 and MATH-500, with some models showing strong general mathematical reasoning but poor extremal-solving skills, and vice versa. This discrepancy highlights a critical gap in current evaluation practices and suggests that existing benchmarks may not comprehensively capture the full spectrum of mathematical reasoning abilities. 

Our contributions and findings are summarized as follows: \textbf{\textit{(i)}}~We construct \texttt{ExtremBench}, a dataset of $93$ extremal problems from inequality proof problems used in Chinese Mathematical Olympiad. While current RL training relies heavily on verifiable answers, we introduce a perspective where we can convert hard-to-verify math proofs into numerically verifiable format, enabling systematic evaluation of optimization reasoning. \textbf{\textit{(ii)}}~We evaluate diverse state-of-the-art reasoning LLMs across multiple model families and scales, providing the first comprehensive assessment of extremal-solving capabilities in contemporary language models.
\textbf{\textit{(iii)}}~Our results demonstrate that LLMs good at general math benchmarks do not always perform well at solving extremal problems, calling for more domain-specific benchmarks to evaluate LLMs' specific mathematical reasoning capabilities.

\section{Related Works}

\paragraph{Mathematical Benchmark Datasets.}
The evaluation of mathematical reasoning capabilities in LLMs has been facilitated by a diverse landscape of benchmark datasets spanning various difficulty levels and mathematical domains. Foundational datasets like GSM8K \citep{cobbe2021training} and MATH-500 \citep{hendrycks2021measuring} established early standards with grade-school and high school competition problems respectively, while MATHQA \citep{amini2019mathqa} provides GRE/GMAT-level multiple-choice questions. For formal theorem proving, miniF2F \citep{zheng2021minif2f} offers a cross-system benchmark with problems from mathematical olympiads. More specialized benchmarks target specific mathematical areas. FIMO \citep{liu2023fimo} focuses on IMO-level algebra and number theory, PutnamBench \citep{tsoukalas2024putnambench} features problems from the prestigious Putnam competition, and ProofNet \citep{azerbayev2023proofnet} addresses undergraduate-level mathematics autoformalization. Recent efforts have pushed toward more challenging problems, with JEEBENCH \citep{arora2023have} covering college-level topics including ODEs and multivariable calculus, MATHBENCH \citep{liu2024mathbench} providing hierarchical coverage from elementary to university mathematics, and GHOSTS \citep{frieder2024mathematical} including graduate-level exercises from advanced mathematics textbooks. To address gaps in specific domains, specialized benchmarks have emerged. CombiBench \citep{liu2025combibench} provides the first comprehensive benchmark for combinatorial mathematics in Lean$4$ with $100$ problems spanning from middle school to IMO level, while HARDMATH \citep{fan2024hardmath} uniquely targets asymptotic analysis and approximation methods with generated problems requiring dominant balance techniques. 

\paragraph{LLM Reasoning for Math}
The capacity of LLMs to perform mathematical reasoning was significantly unlocked by CoT prompting, which elicits intermediate reasoning steps to guide the model toward a solution \citep{openai2024openaio1card, wei2022chain, yang2025qwen3}. Subsequent work enhanced this process through techniques like self-consistency~\citep{wang2022self}, which samples multiple reasoning paths and selects the most frequent answer, mitigating errors from greedy decoding. Progress in the field has been largely driven and measured by performance on a hierarchy of benchmarks, from grade-school word problems such as GSM8K~\citep{cobbe2021training} to challenging competition-level mathematics found in the MATH-500 dataset~\citep{hendrycks2021measuring} and the American Invitational Mathematics Examination~(AIME). A burgeoning area of research now applies LLMs to optimization, primarily focusing on translating natural language problem descriptions into formal models for external solvers, a task evaluated by benchmarks like OptiBench \citep{wang2024optibench} and addressed by frameworks such as LLMOPT \citep{jiang2024llmopt}. However, this paradigm of problem \textit{formulation} for numerical solvers is distinct from the symbolic and logical reasoning required to \textit{solve} mathematical optimization problems directly. As our work highlights, existing benchmarks, while broad, do not systematically evaluate an LLM's intrinsic ability to find extrema under constraints through symbolic manipulation. This leaves a critical gap in understanding a fundamental component of mathematical intelligence, which our proposed \texttt{ExtremBench} aims to address.

\section{The \texttt{ExtremBench} Dataset}

Our dataset construction converts inequality proof problems from mathematical olympiad exercises into standardized extremal problems. We use Anthropic's Claude Opus 4.1 to execute our construction. We describe our methodology below. 

\paragraph{Source Material.} We sourced our problems from \textit{An Introduction to the Proving of Elementary Inequalities}~\citep{han2011introduction}, a popular collection of CMO exercises that contains challenging inequality problems. Given that the original problems were in Chinese, we first employed an LLM to translate them into English while preserving mathematical notation and logical structure. This translation step was verified by bilingual authors to ensure mathematical accuracy and clarity were maintained.

\paragraph{Transformation to Extremal Problems.} For each inequality problem of the form ``prove that $A \leq B$'' or ``prove that $A \geq B$'' under given constraints, we reformulated it as an optimization task: ``find the maximum/minimum of $A - B$'' subject to the same constraints. Specifically, we employed the following prompt for automated conversion:
\begin{quote}
\small
\texttt{Rewrite the inequality proof problem $A \le B$ or $A \ge B$ as an extremum problem: with all original conditions unchanged, reformulate it as "find the extremum of $A-B$." Move all constant terms to the other side so that the objective $A-B$ includes only variable-dependent terms. Output **only** the converted problem inside a single LaTeX code block, with no additional text or explanations.}
\end{quote}
Each converted problem passes rigorous manual verification by the authors to ensure: (1) the transformation correctly preserved all constraints from the original problem, (2) the extremal formulation was mathematically equivalent to the original inequality, and (3) the problem statement was unambiguous and self-contained. From the initial $100$ problems, we retained $93$ after filtering out strict inequalities where equality cannot be achieved. 

\paragraph{The Final Dataset.} Our final \texttt{ExtremBench} dataset comprises $93$ extremal problems, each formulated as a well-defined optimization task asking for either a maximum or minimum value, along with its corresponding  numerical answer. Each problem in the dataset includes clear constraint specifications and an objective function exactly derived from the original inequality. Our dataset includes $62$ minimization problems and $31$ maximization problems. Examples of both the original inequality problems and their converted extremal counterparts are illustrated in Figure~\ref{fig:teaser} and Appendix~\ref{app:example}.

\section{Experiments}

\paragraph{Experimental Setup.}
We evaluate a set of state-of-the-art open-source reasoning LLMs across three model families to assess their extremal-solving capabilities on \texttt{ExtremBench}, including Qwen3 (1.7B, 4B-Thinking-2507, 8B, 14B, 32B, 30B-A3B-Thinking-2507, 235B-A22B-Thinking-2507), GPT-OSS (20B, 120B), and DeepSeek-R1 (1.5B, 7B, 8B, 14B). We use SGLang~\cite{zheng2024sglangefficientexecutionstructured} on NVIDIA B200 GPUs for efficient inference. We report average performance of $3$ repeated trials.

\begin{wrapfigure}{r}{0.5\textwidth}
    \centering
    \vspace{-8pt}
    \includegraphics[width=0.5\textwidth]{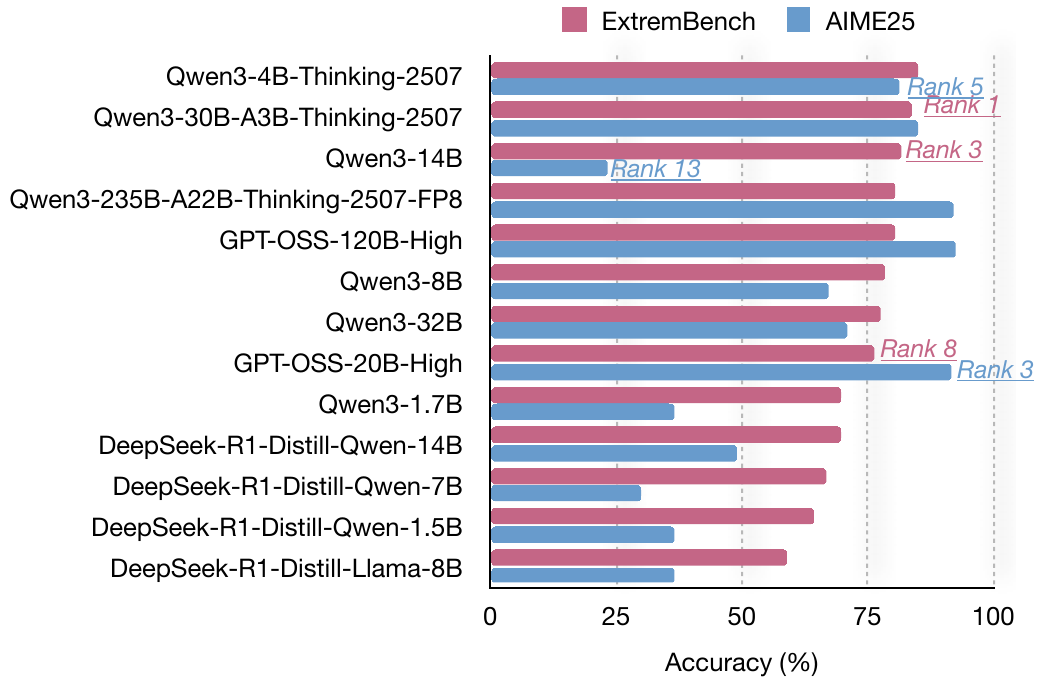}
    \vspace{-22pt}
    \caption{Evaluation results on \texttt{ExtremBench} and AIME25, sorted by the accuracy of \texttt{ExtremBench}.}
    \vspace{-12pt}
    \label{fig:results}
\end{wrapfigure}
\paragraph{Benchmark Results.}  Figure~\ref{fig:results} reveals surprising disparities between extremal-solving and general mathematical reasoning capabilities across model families. We draw several findings: 
(1) While GPT-OSS-120B-High and GPT-OSS-20B-High achieve near-perfect scores on AIME25 ($>90\%$), their \texttt{ExtremBench} performance plateaus around $70\%$, suggesting that strong general mathematical reasoning does not guarantee proficiency in optimization tasks. 
(2) Larger models do not consistently outperform smaller ones on \texttt{ExtremBench}, \textit{e.g.} Qwen3-14B matches the performance of Qwen3-235B despite having $17\times$ fewer parameters, indicating that extremal-solving ability may depend more on specific training data or architectural choices than raw scale. 
(3) The Qwen3-Thinking variants demonstrate the strongest \texttt{ExtremBench} performance ($75-80\%$) despite moderate AIME25 scores, while DeepSeek-R1 models show consistently lower performance on both benchmarks ($50-60\%$ on \texttt{ExtremBench}, $30-40\%$ on AIME25).
These results underscore that extremal problem-solving represents a distinct mathematical competency that current benchmarks fail to capture, highlighting the necessity of specialized evaluation frameworks like \texttt{ExtremBench} for comprehensive assessment of LLM mathematical capabilities.

\section{Conclusion}
In this work, we introduced \texttt{ExtremBench}, a specialized benchmark for evaluating LLMs' extremal-solving capabilities, addressing a critical gap in the benchmark of mathematical reasoning abilities. By transforming inequality proof problems from Chinese Mathematical Olympiad exercise into standardized, numerically verifiable extrema-finding tasks, we provide the first systematic evaluation framework for extremal optimization reasoning, which is a fundamental skill underlying applications in planning, control, resource allocation, and prompt search. Our extensive evaluation across state-of-the-art model families including Qwen3, GPT-OSS, and DeepSeek reveals a suprising insight: performance on extremal problems does not necessarily correlate with success on established mathematical benchmarks like AIME25 and MATH-500, with some models excelling at general mathematical reasoning while struggling with optimization tasks, and others showing the opposite pattern. This discrepancy exposes a significant blind spot in current evaluation practices and suggests that existing benchmarks fail to capture the full spectrum of mathematical intelligence. Our findings call for a more nuanced approach to benchmarking LLM capabilities, emphasizing the need for domain-specific evaluations that can identify distinct reasoning competencies beyond general mathematical problem-solving. 

\textbf{Future work} could explore several directions: 
(1)~Our transformation methodology from hard-to-verify inequality proofs to numerically verifiable extremal problems demonstrates a novel paradigm that could be extended to other mathematical domains. This approach could potentially unlock vast repositories of proof-based problems for RL training, as verifiable answers are crucial for test-time scaling and reward modeling. We envision similar transformations for problems in combinatorics~(converting existence proofs to counting problems), geometry~(converting congruence proofs to distance optimization), and analysis~(converting convergence proofs to rate-of-convergence optimization).
(ii) Expanding ExtremBench to include multi-objective optimization, constrained optimization with equality constraints, and discrete optimization problems would provide a more comprehensive evaluation of LLMs' optimization reasoning capabilities across different mathematical structures.
(2)~Investigating the underlying mechanisms behind the observed discrepancy between general mathematical reasoning and extremal-solving abilities could reveal fundamental insights about how LLMs encode and apply different types of mathematical knowledge, potentially informing more targeted training strategies.
(3)~Developing specialized supervised fine-tuning or RL approaches that specifically target extremal reasoning, possibly through curriculum learning that progressively increases constraint complexity and dimensionality.
(4)~Exploring whether the extremal-solving capability serves as a better predictor for downstream tasks in scientific computing, operations research, and automated theorem proving, where optimization reasoning is fundamental.

\section*{Acknowledgment}
We thank Pingzhi Li for thoughtful discussions and feedback during this work.

\bibliographystyle{abbrv}
\bibliography{reference}

\appendix

\section{Examples in \texttt{ExtremBench}}~\label{app:example}

Below we present representative examples from \texttt{ExtremBench}:

\textbf{Example 1:} For $a, b, c, d \in \mathbb{R}^+$ with $abcd = 1$, find the minimum of
$$\frac{1}{(1 + a)^2} + \frac{1}{(1 + b)^2} + \frac{1}{(1 + c)^2} + \frac{1}{(1 + d)^2} $$
\textbf{Answer:} $1$

\textbf{Example 2:} For $a, b, c, d > 0$ with $a + b + c + d = 1$, find the minimum of
$$(1 - \sqrt{a})(1 - \sqrt{b})(1 - \sqrt{c})(1 - \sqrt{d}) - \sqrt{abcd}$$
\textbf{Answer:} $0$

\textbf{Example 3:} For $a, b, c > 0$ with $a + b + c = 3$, find the minimum of
$$\sqrt{3 - bc} + \sqrt{3 - ca} + \sqrt{3 - ab}$$
\textbf{Answer:} $3\sqrt{2}$

\textbf{Example 4:} For $a, b, c \in [0, 1]$, find the maximum of
$$(a + b + c)\left(\frac{1}{bc + 1} + \frac{1}{ca + 1} + \frac{1}{ab + 1}\right)$$
\textbf{Answer:} $5$

\textbf{Example 5:} For $x, y, z > 0$, find the minimum of
$$\sum_{cyc} \sqrt{\frac{(y + z)^2yz}{(x + y)(x + z)}} - \sum x$$
\textbf{Answer:} $0$










\end{document}